\documentclass{article}
\usepackage{spconf,amsmath,graphicx}
\usepackage[nolist]{acronym}
\usepackage{booktabs} 
\usepackage{siunitx}
\usepackage{xurl}
\usepackage{subcaption}
\usepackage{xcolor}
\usepackage{hyperref}
\usepackage{microtype} 

\begin{acronym}
\acro{mde}[MDE]{Mean Distance Error}
\acro{caffe}[CaFFe]{``CAlving Fronts and where to Find thEm''}
\acro{sar}[SAR]{Synthetic Aperture Radar}
\acro{s1}[S1]{Sentinel-1}
\acro{iou}[IoU]{Intersection over Union}
\acro{gru}[GRU]{Gated Recurrent Unit}
\end{acronym}

\title{Few-Shot Domain Adaptation with Temporal References and Static Priors for Glacier Calving Front Delineation}

\name{
\begin{tabular}{c}
\em Marcel Dreier$^{1,\star}$ \qquad Nora Gourmelon$^{1,\star}$ \qquad Dakota Pyles$^2$ \qquad Thorsten Seehaus$^2$\\
\em Matthias H. Braun$^2$ \qquad Andreas Maier$^1$ \qquad Vincent Christlein$^1$
\end{tabular}
}
\address{$^{\star}$Shared first-authorship \\
$^{1}$Pattern Recognition Lab,
Friedrich-Alexander-Universität Erlangen-Nürnberg, Erlangen, Germany \\
$^{2}$Institut für Geographie, Friedrich-Alexander-Universität Erlangen-Nürnberg, Erlangen, Germany}

\begin{document}
\maketitle
\begin{abstract}
During benchmarking, the state-of-the-art model for glacier calving front delineation achieves near-human performance. However, when applied in a real-world setting at a novel study site, its delineation accuracy is insufficient for calving front products intended for further scientific analyses. This site represents an out-of-distribution domain for a model trained solely on the benchmark dataset. By employing a few-shot domain adaptation strategy, incorporating spatial static prior knowledge, and including summer reference images in the input time series, the delineation error is reduced from \SI{1131.6}{\metre} to \SI{68.7}{\metre} without any architectural modifications. These methodological advancements establish a framework for applying deep learning-based calving front segmentation to novel study sites, enabling calving front monitoring on a global scale.
\end{abstract}
\section{Introduction}
\label{sec:intro}

With accelerating climate change manifesting through increasingly frequent floods, wildfires, and other extreme events~\cite{IPCC.2021}, monitoring climate indicators such as glaciers remains essential. 
For marine-terminating glaciers, changes in the position of the calving front, defined as the boundary of the glacier that faces the ocean, provide critical information on glacier dynamics. 
\ac{sar} enables year-round observations independent of weather and illumination conditions, which is particularly important in polar regions.
On the \ac{caffe} benchmark dataset~\cite{Gourmelon.2022}, calving front delineation performance close to human accuracy has already been achieved~\cite{Dreier.2025}. 
However, transferring these results from a benchmarking setting to real-world applications exposes important limitations. 
New scenarios not covered by the benchmark can arise, including data from new sensors, different acquisition geometries, alternative polarization configurations, and study sites beyond the seven glaciers included in the benchmark, which are located in Alaska, Greenland, and the Antarctic Peninsula.
In this paper, we use the entire Svalbard archipelago as a case study. 
The current state-of-the-art model, Tyrion-T-GRU, trained solely on \ac{caffe}, reaches an error of \SI{1131.6}{\metre}, which is not sufficient for calving front products intended for further scientific analyses.

Tyrion-T-GRU processes a time series of \ac{sar} acquisitions and produces semantic segmentation for each acquisition into glacier, ocean, rock, and areas with no available information (NA) like \ac{sar} shadows~\cite{Dreier.2025}. The calving fronts are subsequently derived from the segmentation maps. 
Glaciers in Svalbard represent an out-of-distribution domain relative to the benchmark dataset~\cite{Gourmelon.2022}, with differing geometries, surface conditions, and climatic influences. To reduce this domain shift, we compile a new dataset that includes one manual annotation per glacier in Svalbard. These labels are used for additional training alongside \ac{caffe}.
The test set for the Svalbard study site comprises acquisitions from three years prior to the training set.
In contrast to Marochov et al.~\cite{Marochov.2021}, who integrated multiple images per glacier, we demonstrate that a single label per glacier provides sufficient guidance for adaptation.

Ice mélange, which is a mixture of sea ice and icebergs, poses a further challenge for \ac{sar}-based segmentation because its backscattering characteristics are similar to those of glacial ice~\cite{Baumhoer.2018}. 
Tyrion-T-GRU~\cite{Dreier.2025} reduces this ambiguity by sharing information across a sequence of eight images, not all of which necessarily contain ice mélange. 
Persistent segmentation errors remain, particularly when ice mélange dominates each scene in the time series. To improve performance under these conditions, we introduce an inference strategy based on reference summer acquisitions, when ice mélange is typically absent. The time series therefore consists of an annual composite of images rather than strictly consecutive acquisitions.

Finally, we incorporate a static rock mask as an additional input modality. Rock locations remain stable over the considered temporal range and encode prior knowledge on local glacier geometry. This information supports the discrimination between rock, glacier ice, and ice mélange in the segmentation process.

Our contributions are:
\begin{itemize}
\item A few-shot domain adaptation strategy that reduces the domain shift between the benchmark dataset and Svalbard by using only one manual label per glacier in Svalbard.
\item The integration of summer reference images to improve performance on ice mélange-affected scenes.
\item A multi-modal training approach that incorporates static rock masks to provide additional spatial prior information.
\end{itemize}
With these three techniques, we are able to reduce the \ac{mde} from \SI{1131.6}{\metre} to \SI{68.7}{\metre}.

\begin{figure}[t]
    \centering
    \includegraphics[width=\linewidth]{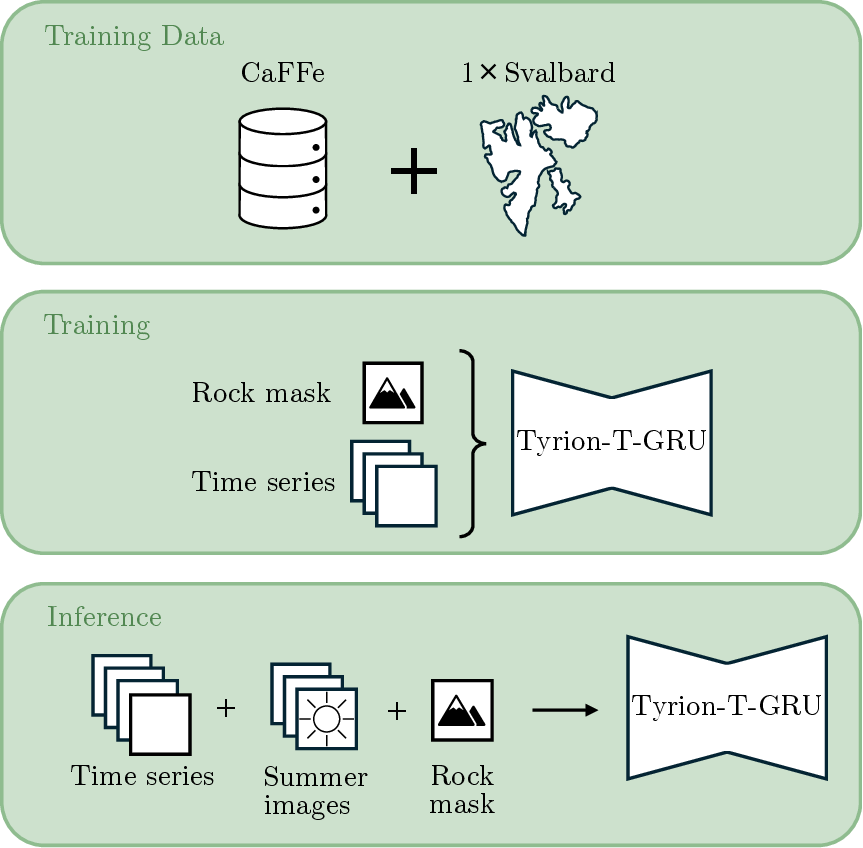}
    \caption{Three methodological advancements.}
    \label{fig:summary}
\end{figure}

\section{Background}
This work builds on the Tyrion-T-GRU model introduced by Dreier et al.~\cite{Dreier.2025} and the \ac{caffe} benchmark dataset~\cite{Gourmelon.2022}. Both components are briefly summarized in the following sections.

    \subsection{Tyrion-T-GRU} 
    Tyrion-T-GRU~\cite{Dreier.2025} combines an ImageNet-pretrained~\cite{deng2009imagenet} SwinV2 encoder~\cite{liu2022swin} with a convolutional decoder composed of ResBlocks and UpsampleBlocks. Standard skip connections link encoder and decoder. Temporal information is introduced through bidirectional \acp{gru}~\cite{gru} inserted after the last three SwinBlocks in the encoder, and through one-dimensional temporal convolutions placed after the first three ResBlocks in the decoder.
    The model processes time series of eight $512 \times 512$ \ac{sar} images and produces a segmentation map of identical size for each time step, following a multi-temporal strategy that preserves temporal resolution throughout the sequence. In contrast, the only other temporal model for calving-front delineation~\cite{zhao_cisnet_2025} adopts a mono-temporal approach that outputs a single aggregated segmentation map per time series. From the $512 \times 512$ predictions, only the central $256 \times 256$ region is retained to ensure sufficient spatial context.

    \subsection{Benchmark Dataset}
    The \ac{caffe} benchmark dataset~\cite{Gourmelon.2022} consists of \ac{sar} satellite images of seven marine-terminating glaciers located on the Antarctic Peninsula, Greenland, and Alaska. 
    The data were acquired by multiple satellite missions, including \ac{s1}.
    The images are single-channel intensity data in either HH or VV polarization.
    They were captured between 1996 and 2020 and have spatial resolutions ranging from \SIrange{7}{20}{\metre} per pixel.
    We use both the training and test splits for training, giving a total of 681 images, because the objective is not to benchmark on \ac{caffe} but to generalize to the Svalbard study region.
    Each \ac{sar} image includes two labels: a zone label that provides a segmentation into glacier, ocean, rock, and no information available (NA), and a direct calving front label. The zone label is used for training the model, while the front label is used for evaluation after post-processing.
    Three metrics are employed for evaluation. The \ac{iou} is used for the intermediate zone segmentation result. The \ac{mde}~\cite{Gourmelon.2022} quantifies the distance in meters between predicted and reference calving fronts. As the predicted segmentation maps sometimes do not show a boundary between ocean and glacier, such that no calving front can be extracted during post-processing, the final metric counts the number of images with no predicted calving front.

\begin{figure*}
    \centering
    \begin{subfigure}{0.16\linewidth}
        \includegraphics[width=\textwidth]{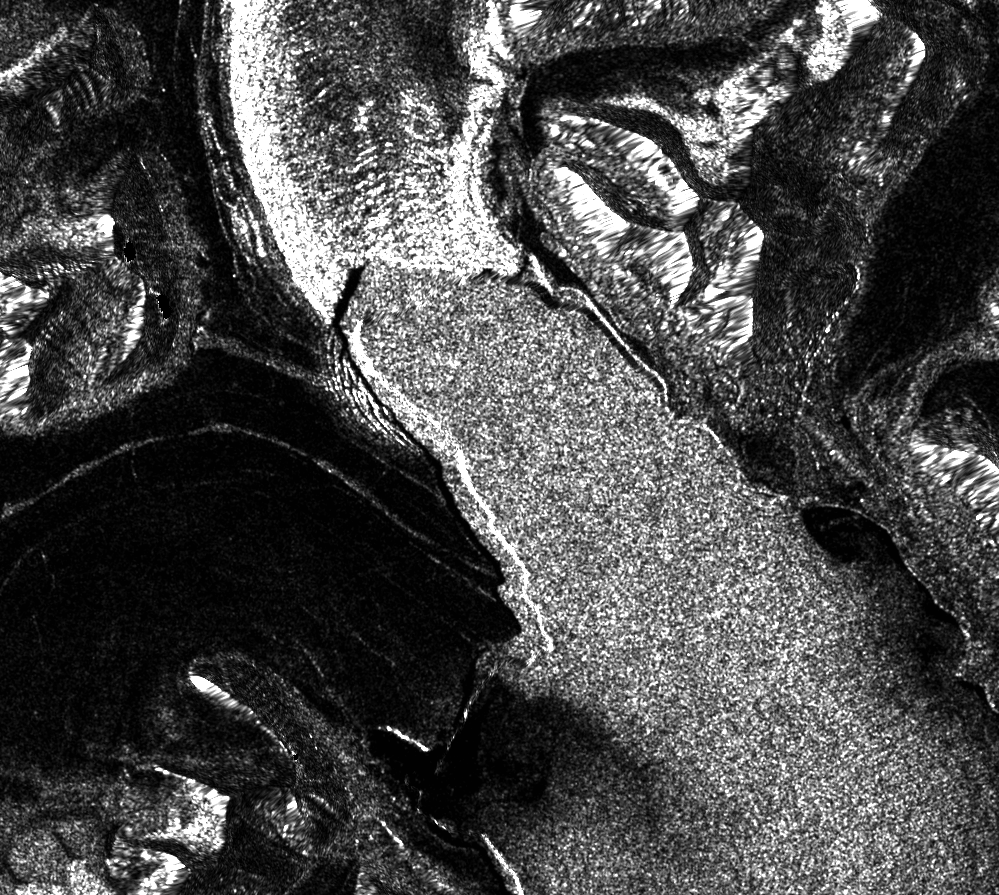}
        \caption{\ac{sar} Image}
        \label{fig:sar}
    \end{subfigure}
    \hfill
    \begin{subfigure}{0.16\linewidth}
        \includegraphics[width=\textwidth]{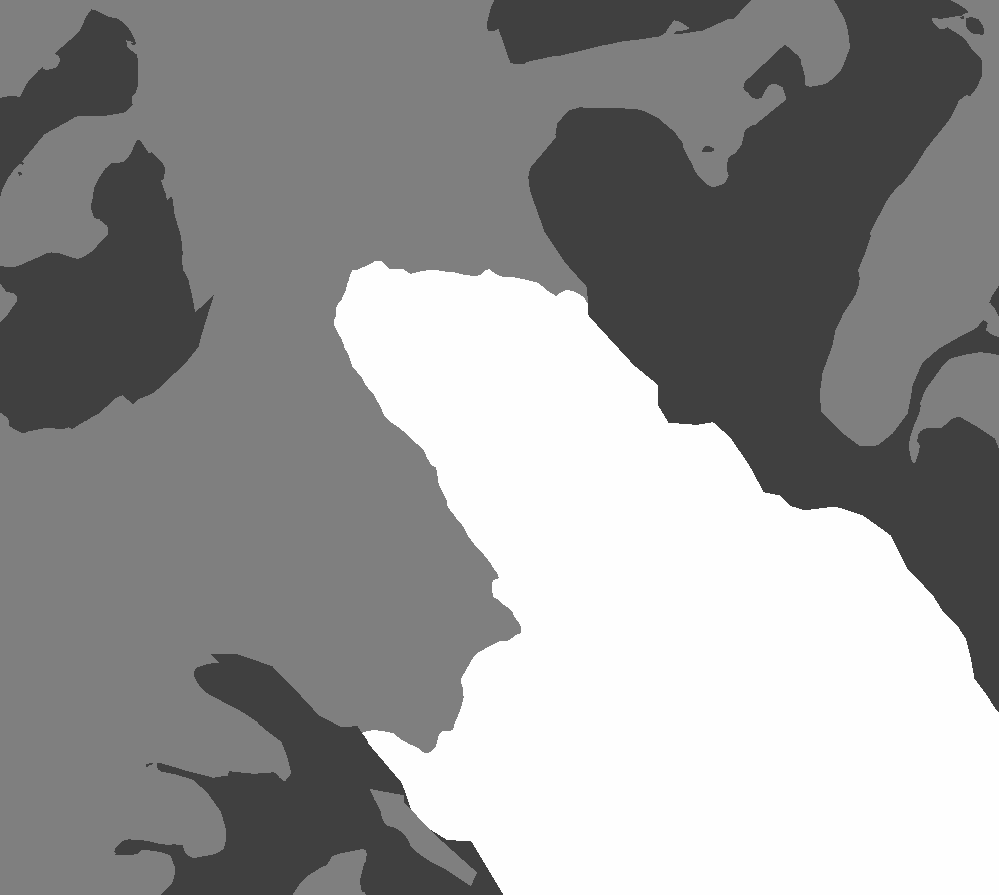}
        \caption{Ground Truth}
        \label{fig:ground_truth}
    \end{subfigure}
    \hfill
    \begin{subfigure}{0.16\linewidth}
        \includegraphics[width=\textwidth]{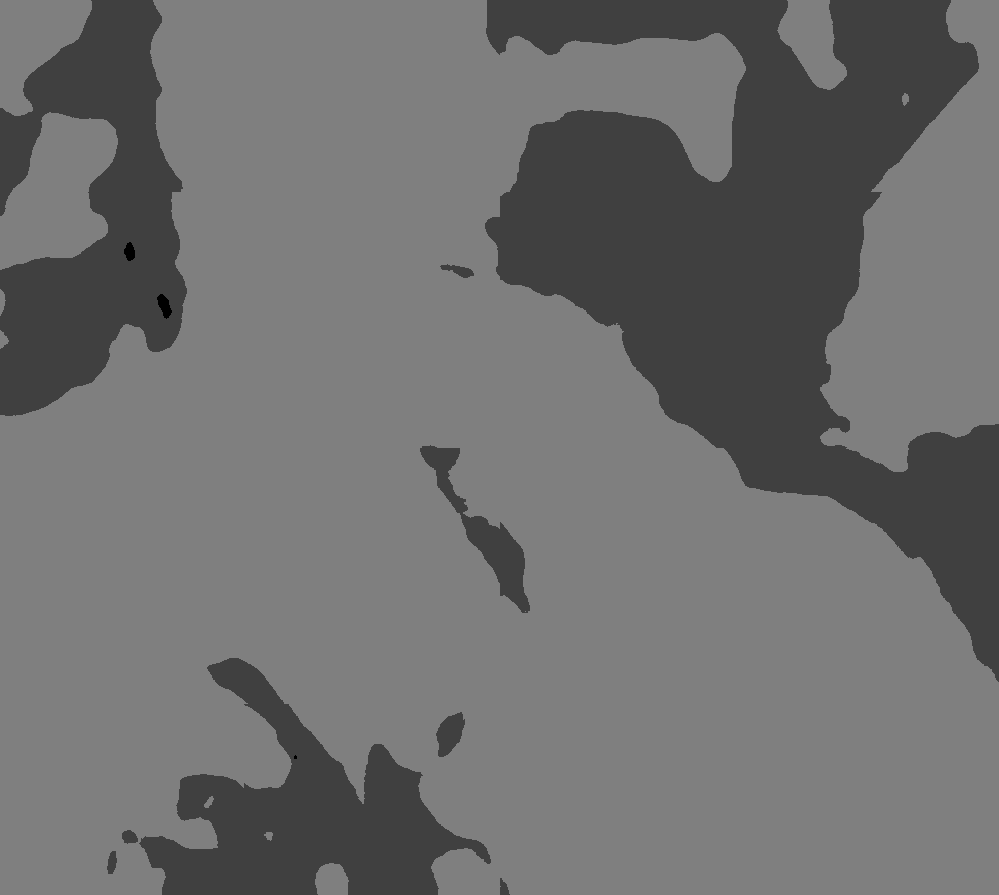}
        \caption{Tyrion-T-GRU}
        \label{fig:tyrion-t-gru}
    \end{subfigure}
    \hfill
    \begin{subfigure}{0.16\linewidth}
        \includegraphics[width=\textwidth]{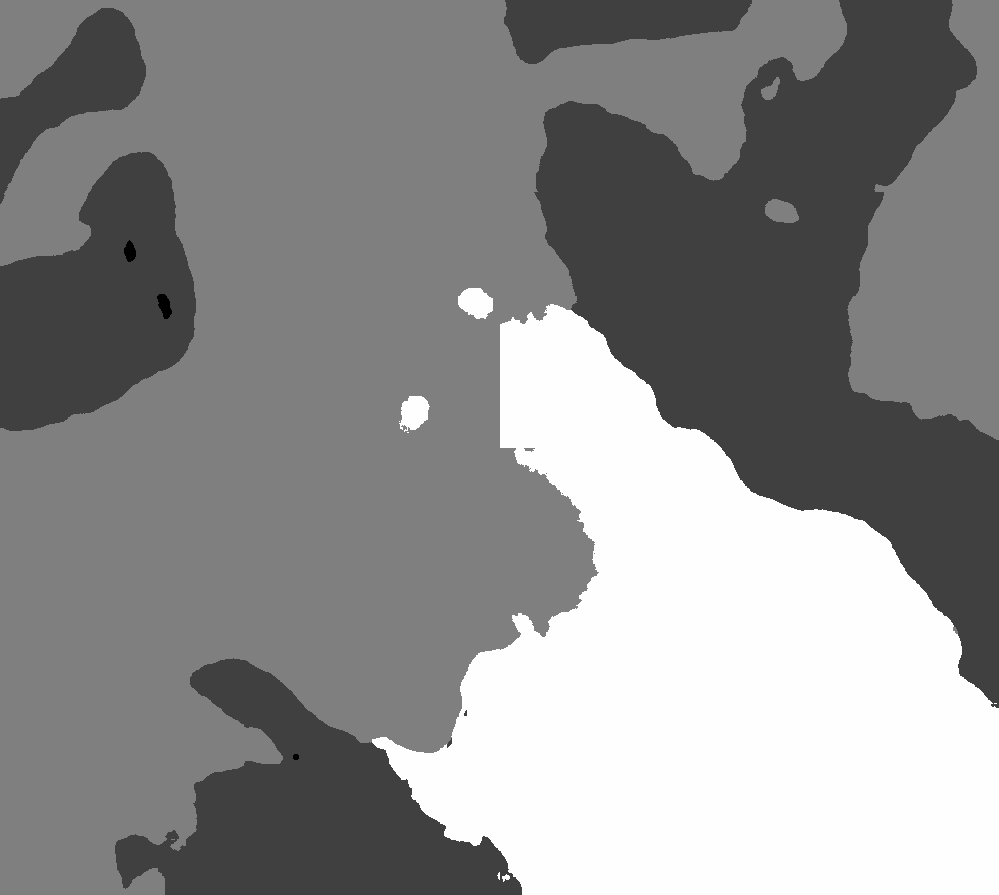}
        \caption{+ Few-Shot}
        \label{fig:few_shot}
    \end{subfigure}
    \hfill
    \begin{subfigure}{0.16\linewidth}
        \includegraphics[width=\textwidth]{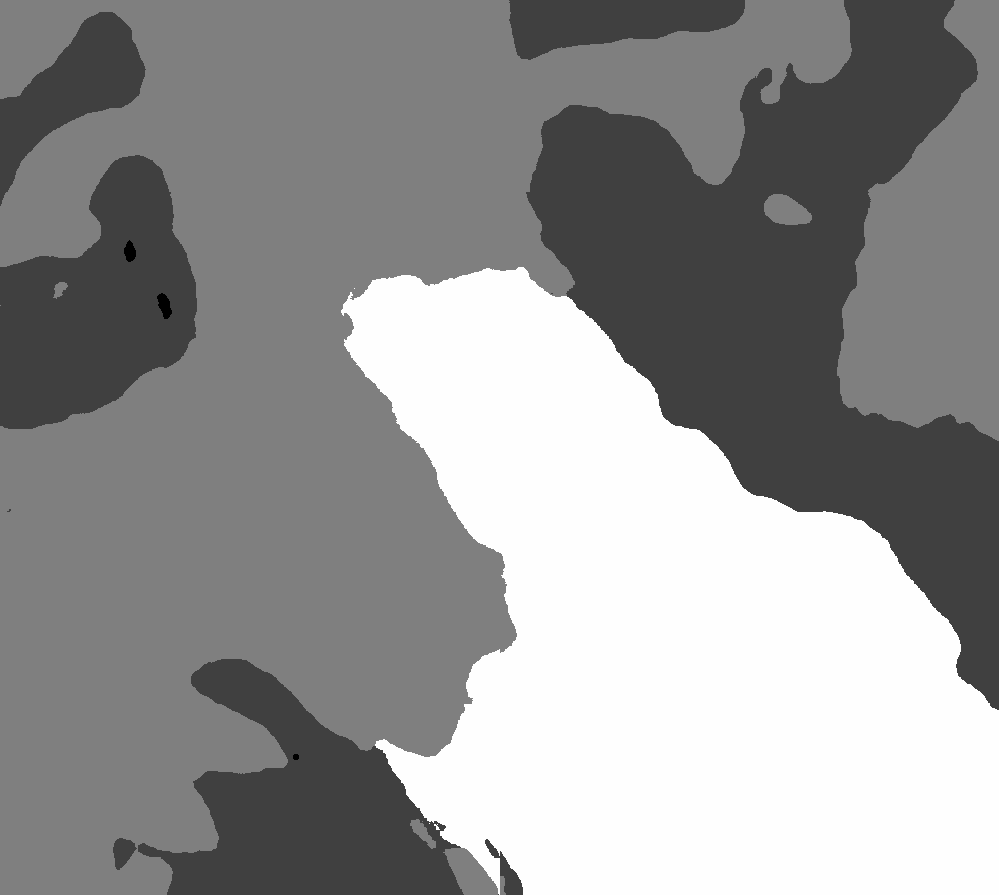}
        \caption{+ Summer Ref.}
        \label{fig:summer_ref}
    \end{subfigure}
    \hfill
    \begin{subfigure}{0.16\linewidth}
        \includegraphics[width=\textwidth]{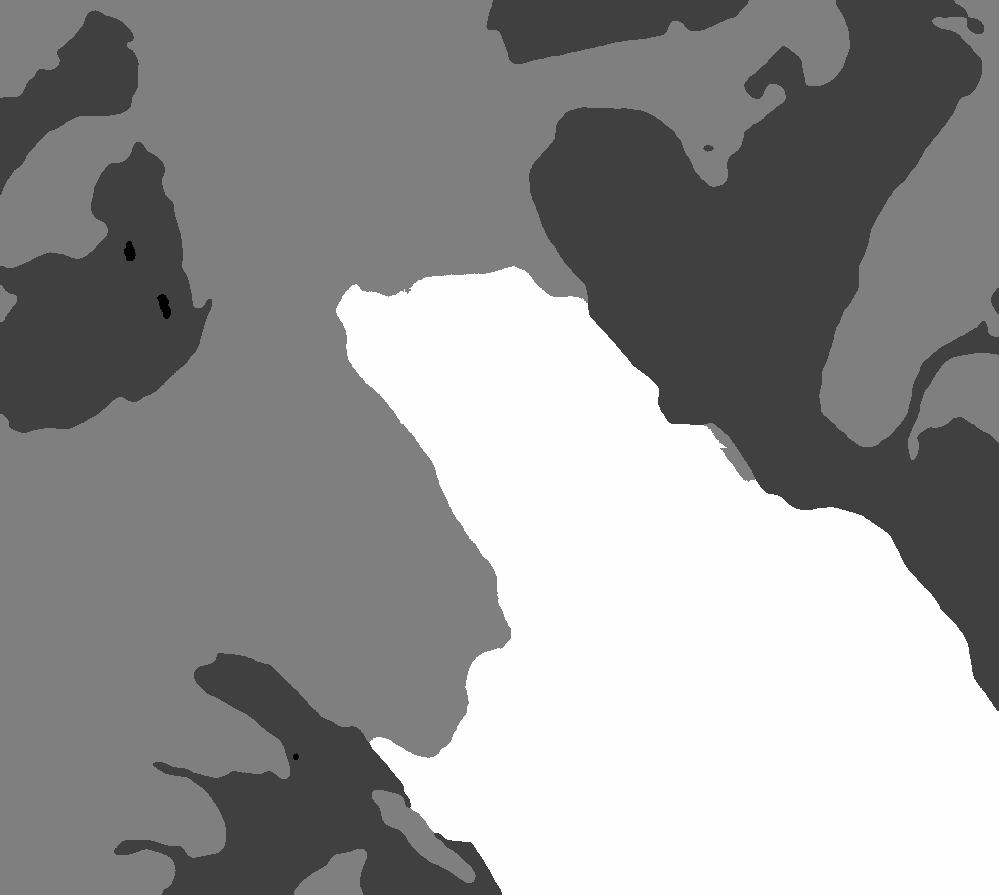}
        \caption{+ Rock Mask}
        \label{fig:rock_mask}
    \end{subfigure}
    \caption{Segmentation results for all experiments. Black corresponds to NA, dark gray to rock, light gray to glacier, and white to ocean and ice mélange.}
    \label{fig:zones}
\end{figure*}

\section{Methodology}
An overview of our methodological advancements can be seen in Fig.~\ref{fig:summary}. The three advancements are described in detail in the following sections.

    \subsection{Few-shot Domain Adaptation} 
    For data-efficient domain adaptation, we assembled a \acl{s1} dataset that includes one manual calving front delineation per glacier in Svalbard, all acquired in summer 2019. 
    This results in 145 annotations with corresponding images.
    Analogous to \ac{caffe}, these annotations are used to construct a front label and a zone label, which are employed for evaluation and training, respectively.
    To enable the training of a temporal model, we collected all \ac{s1} acquisitions from July and August 2019. 
    Although the calving front may physically move over the two-month period, the resulting positional shift in the image is minimal, and we therefore use the same manual annotation as label for all corresponding images.
    For validation, ten manual annotations were produced for each of four glaciers using \ac{s1} images acquired in 2016. For all remaining \ac{s1} images from 2016, the temporally closest manual annotation is used to construct dense time series. The test set is compiled in the same manner for an additional set of five glaciers. 
    Metric computation is restricted to outputs that correspond directly to available manual labels. 
    The final dataset comprises 5539 training images, 192 validation images, and 228 test images.
    In contrast to \ac{caffe}, which has varying resolutions and includes only HH and VV polarization, all images have a spatial resolution of \SI{10}{\metre} per pixel and are single-channel intensity data with HH, HV, VV, or VH polarization.
    Tyrion-T-GRU is trained jointly on this constructed dataset and \ac{caffe}, and the \ac{iou} on the Svalbard validation set is employed as the early stopping criterion.

    \subsection{Summer Reference Images}
    When using strictly consecutive acquisitions in a time series, it may happen that all images contain ice mélange in front of the calving front, which significantly impairs segmentation performance. To mitigate this issue and improve performance on ice-mélange-affected images, we introduce a novel inference strategy.
    We augment each time series with three summer reference images, for which the probability of ice mélange is naturally low. 
    In the resulting series, there is a reference image from July, August, and September of the same year.
    The remaining four images form the central subset of the time series used for the analysis.
    
    \subsection{Rock Masks}
    To incorporate readily available prior knowledge, we generate static rock masks for each glacier in Svalbard. 
    We construct a conjunction of glacier outline polygons from the Randolph Glacier Inventory version 6~\cite{RGI.2017} and glacier tongue polygons with accurate, up-to-date calving fronts that were manually mapped using Landsat 8 imagery~\cite{Kochtitzky.2022_glacier_outlines}. 
    The resulting polygons define the glacier area. 
    Rock areas are then identified by assigning all regions within the OpenStreetMap coastline of Svalbard~\cite{OpenStreetMap}, that are not covered by the glacier polygons, to the rock masks. 
    Finally, manual refinements were applied in the vicinity of the calving front using Sentinel-2 imagery.

    One rock mask corresponding to the glacier represented in the time series is added as an additional modality to the time series input, providing spatial prior information that supports calving front localization.

\section{Experimental Setup}
We perform several experiments that build on each other. 
As a baseline, we use Tyrion-T-GRU~\cite{Dreier.2025}, diverging from the benchmarking setup by training the model on the entire \ac{caffe} dataset rather than only the \ac{caffe} training set. 
In the second experiment, referred to as few-shot, the \ac{caffe} dataset is augmented with the Svalbard dataset. 
Next, the summer reference dates are incorporated, and in the final experiment, rock masks are added as an additional input.
An ensemble is subsequently built using five retrained model versions from the final experiment.
Class-wise uncertainties for the ensemble outputs are estimated as the standard deviation of the corresponding class logits across the five models.
All experiments are evaluated on the Svalbard test set.

\begin{table*}[t]
   \begin{minipage}{\textwidth}
    \centering
    \caption{Overview of the resulting evaluation metrics.}
    \label{tab:distance_errors_annotators}
  \begin{tabular}{c|cc|ccccc}
        \toprule
        & \multicolumn{2}{c}{\textit{Calving Front Segmentation}} && \multicolumn{4}{c}{\textit{Zone Segmentation \ac{iou}}} \\
         \textit{Model} & \textit{MDE} $\downarrow$ & \textit{Missing Fronts}$\downarrow$ & \textit{All}$\uparrow$ & \textit{NA} $\uparrow$  & \textit{Rock}$\uparrow$  &  \textit{Glacier}$\uparrow$ & \textit{Ocean}$\uparrow$  \\
        \midrule
         Tyrion-T-GRU  & $1131.6\pm179.3$ & $5.6 \pm 2.7$ & $53.9\pm1.8$ & $23.9\pm2.0$ & $58.3\pm3.4$ & $67.3\pm2.2$ & $66.2\pm6.2$  \\
         + Few-Shot  & $445.3\pm81.0$ & $7.2 \pm 2.9$ & $68.7\pm0.5$ & $27.9\pm0.7$ & $77.4\pm0.6$ &  $82.6\pm1.0$ &  $87.0\pm2.1$ \\
         + Summer Ref. & $204.6\pm57.5$ & 6.2 $\pm$ 1.5 & $72.4\pm0.4$ & $28.4\pm0.6$ & $81.8\pm 0.6$& $87.1\pm0.8$ & $92.5\pm0.5$   \\
         + Rock Mask & $103.6\pm31.4$ & \textbf{0.0} $\pm$ \textbf{0.0} & $80.6\pm0.3$ & $28.4\pm0.4$ & $99.1\pm 0.0$& $97.8\pm0.5$ & $97.0\pm0.7$   \\
         \midrule
         + Ensemble &  $\mathbf{68.7}$ & \textbf{0.0} & $\mathbf{81.1}$ & $\mathbf{28.9}$ & $\mathbf{99.3}$& $\mathbf{98.4}$ & $\mathbf{97.7}$   \\
         \bottomrule
    \end{tabular}
     \end{minipage}
\end{table*}

\begin{figure*}
    \centering
    \begin{subfigure}{0.19\linewidth}
        \includegraphics[width=\textwidth, trim=0 0 0 95, clip]{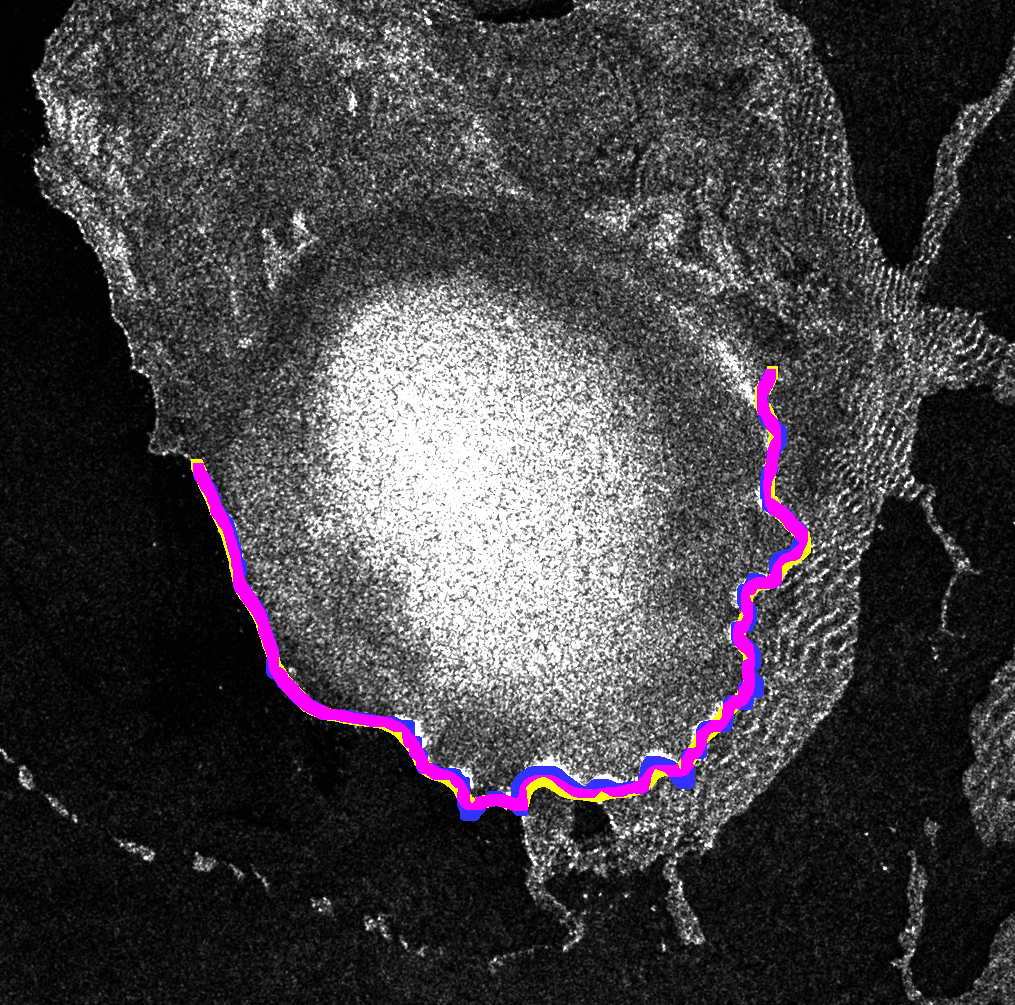}
    \end{subfigure}
    \hfill
    \begin{subfigure}{0.19\linewidth}
        \includegraphics[width=\textwidth]{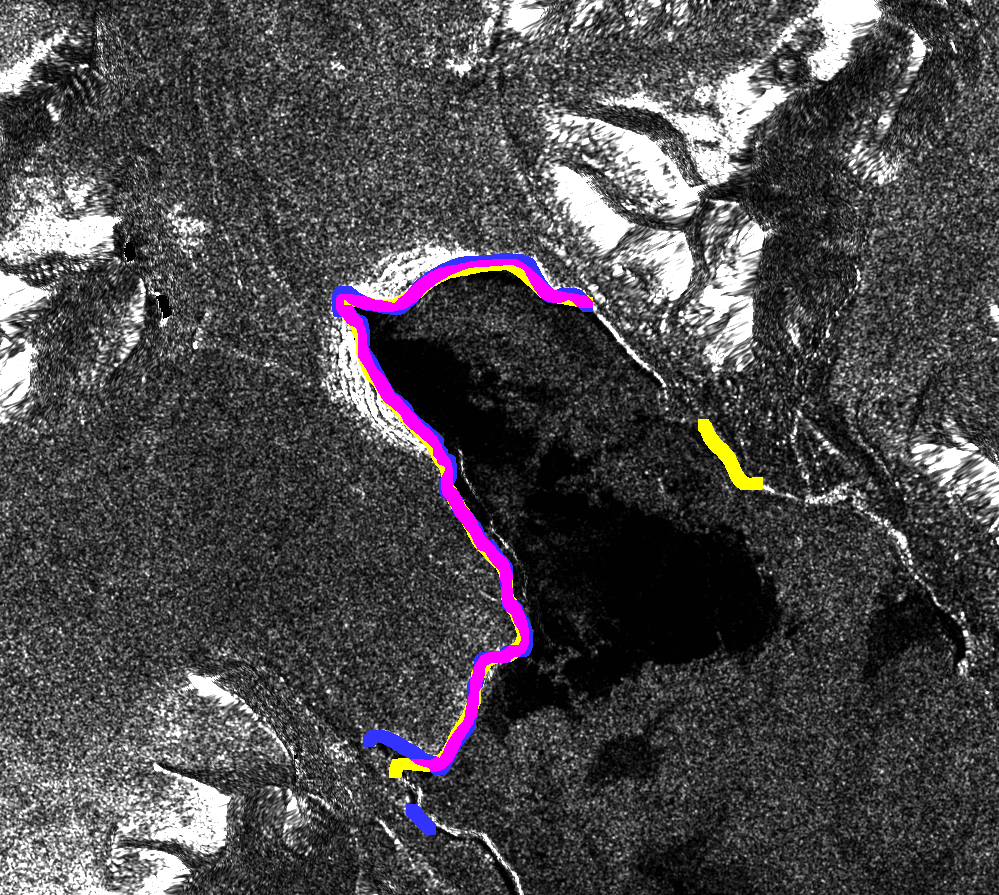}
    \end{subfigure}
    \hfill
    \begin{subfigure}{0.19\linewidth}
        \includegraphics[width=\textwidth, trim=0 12 0 0, clip]{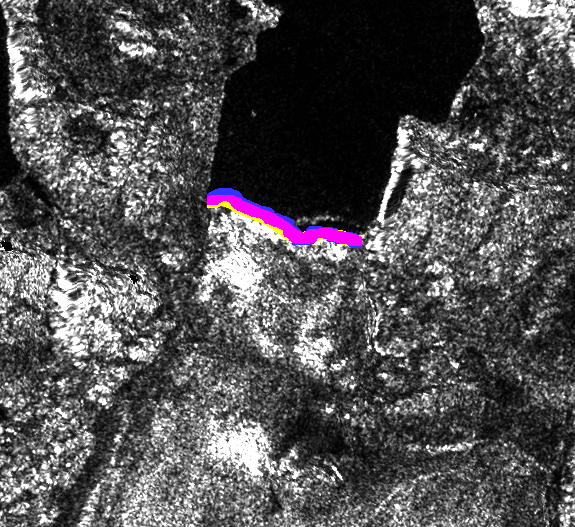}
    \end{subfigure}
    \hfill
    \begin{subfigure}{0.19\linewidth}
        \includegraphics[width=\textwidth, trim=0 0 0 34, clip]{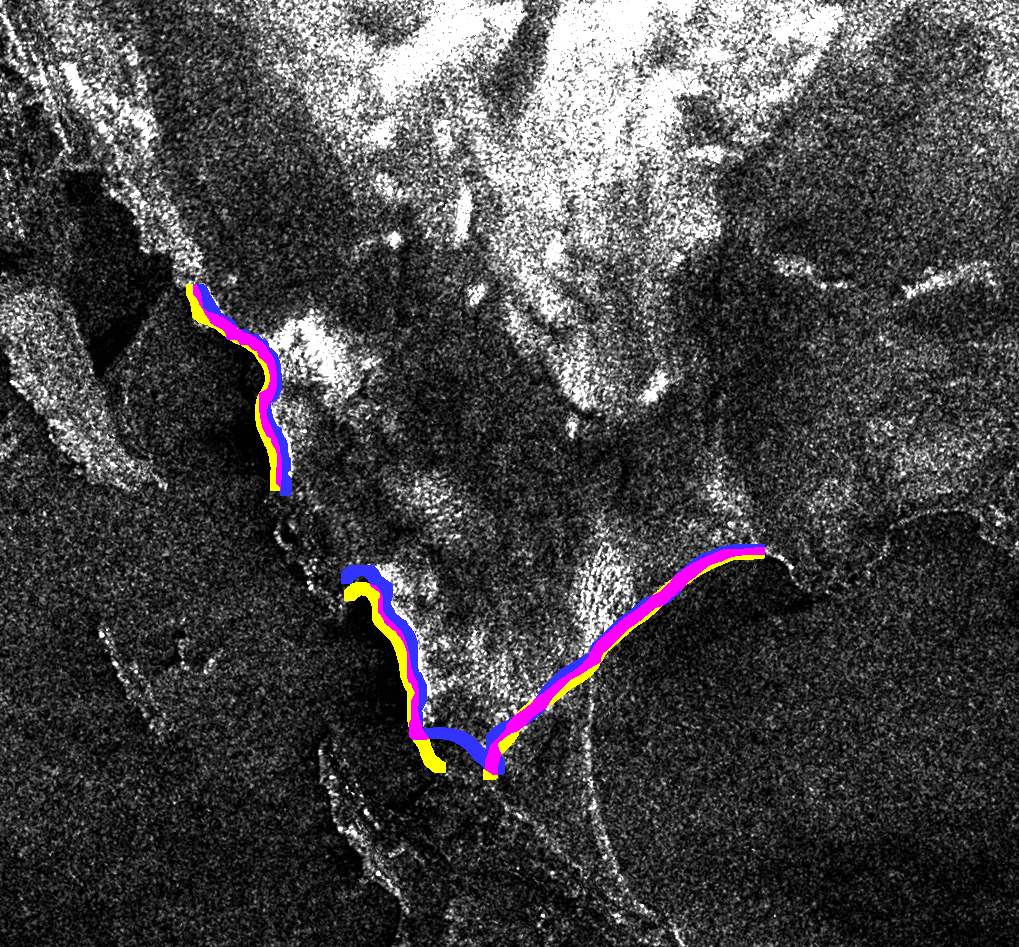}
    \end{subfigure}
    \hfill
    \begin{subfigure}{0.19\linewidth}
        \includegraphics[width=\textwidth, trim=0 142 0 100, clip]{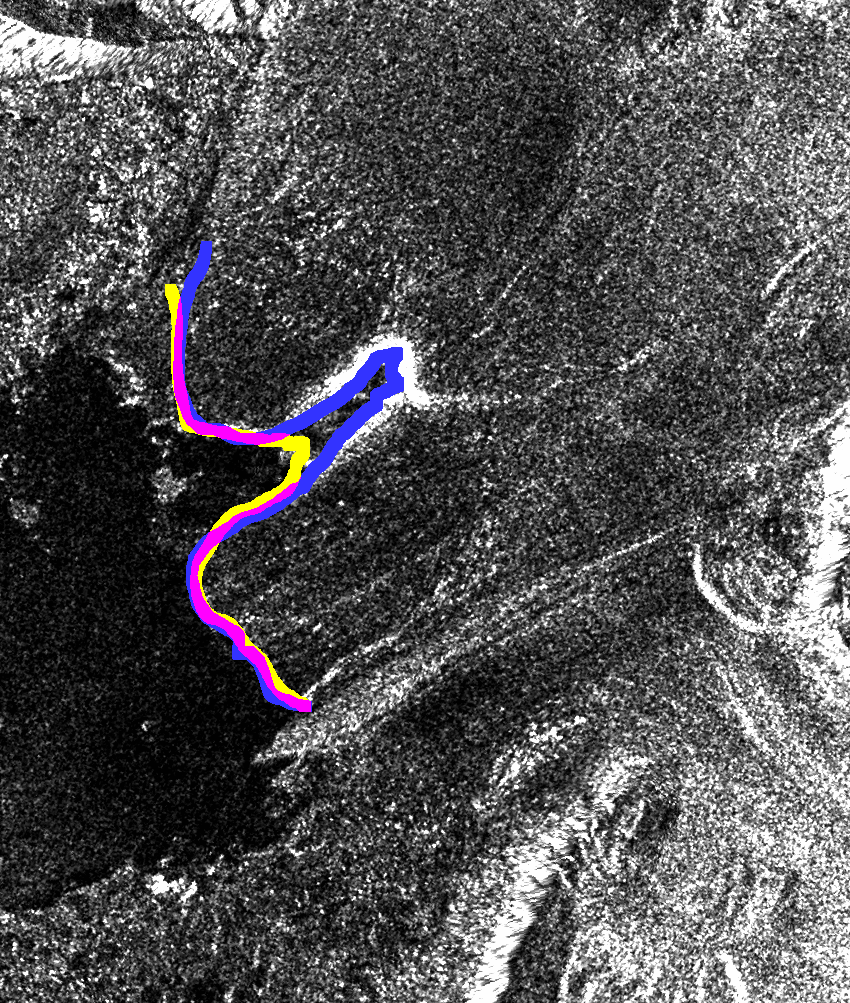}
    \end{subfigure}
    \caption{Ensemble results for all five test set glaciers. Predicted fronts, ground truth fronts, and their overlap are shown in \fboxsep=1pt\colorbox{yellow!100}{yellow}, \fboxsep=1pt\colorbox{blue!100}{\color{white}blue}, and \fboxsep=1pt\colorbox{magenta!100}{\color{white}pink}, respectively. All calving fronts are dilated for visualization purposes.}
    \label{fig:vis}
\end{figure*}

\section{Results and Discussion}
The quantitative results for all experiments are given in Table~\ref{tab:distance_errors_annotators}.
Applying few-shot domain adaptation provides the largest performance gains among the improvements, lowering the \ac{mde} by \SI{686.3}{\metre} and reaching an \ac{iou} of $68.7$, with roughly equal increases in rock, glacier, and ocean \ac{iou}. 
This performance gain can be explained by both the larger training dataset size and the closed domain gap.
Incorporating summer reference images reduces the \ac{mde} by another \SI{240.7}{\metre} and, surprisingly, yields comparable performance gains across all \acp{iou}. 
Consequently, this strategy not only improves the recognition of ice mélange but also benefits the segmentation overall.
Adding rock masks as additional input eliminates missing fronts and halves the prior \ac{mde}. 
Moreover, rock \ac{iou} and glacier \ac{iou} experience substantial increases of $17.3$ and $10.7$, respectively, while ocean \ac{iou} increases by $4.5$.
The sharp increase in rock \ac{iou} was expected, but the additional gains in ocean and glacier \ac{iou}, together with the reduced \ac{mde}, indicate that incorporating spatial prior knowledge is beneficial overall.
Figure~\ref{fig:zones} shows zone segmentation results across the experiments.
The standard deviations of all metrics decrease over the experiments, indicating improved agreement among the retrained model versions.  

\begin{figure*}
    \centering
    \begin{subfigure}{0.19\linewidth}
        \includegraphics[width=\textwidth, trim=0 0 0 95, clip]{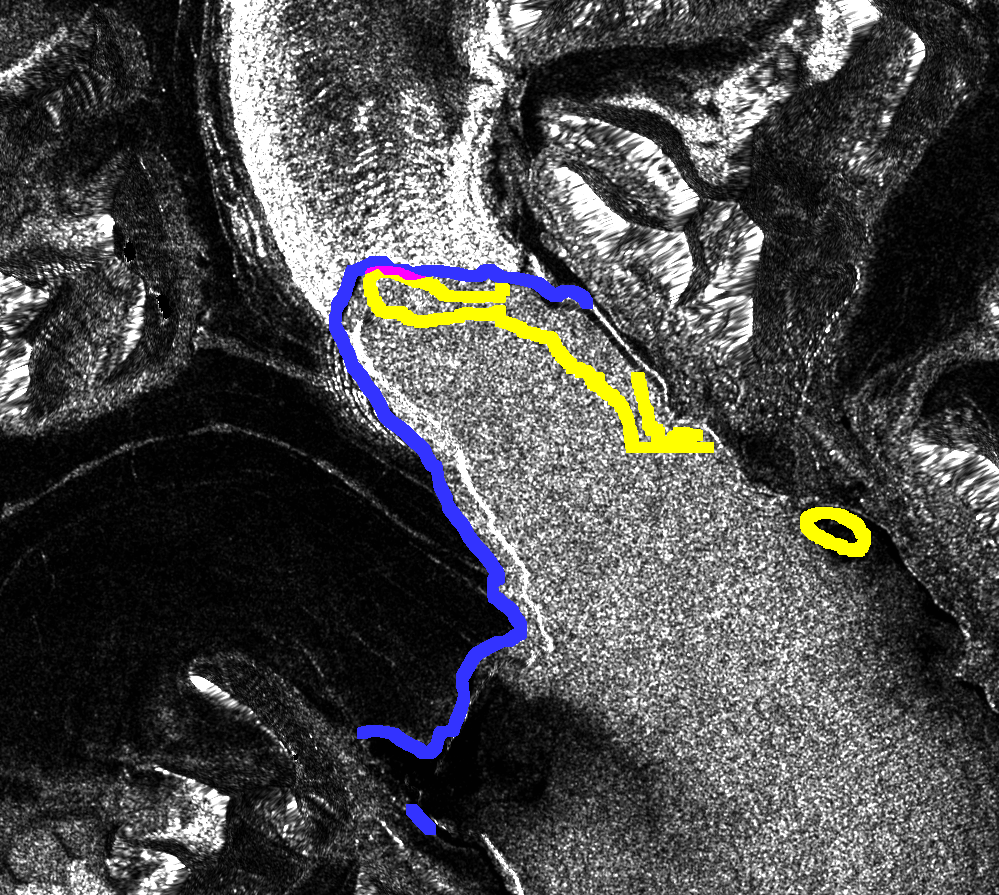}
    \end{subfigure}
    \hfill
    \begin{subfigure}{0.19\linewidth}
        \includegraphics[width=\textwidth, trim=0 0 0 95, clip]{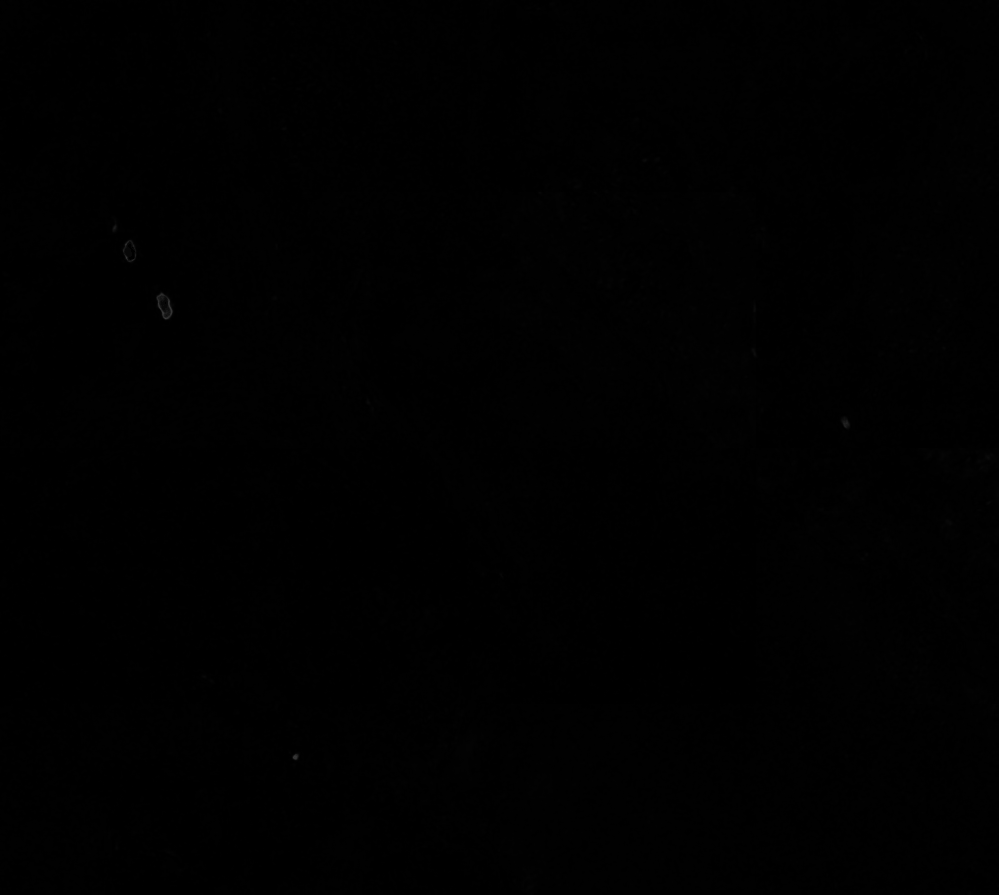}
    \end{subfigure}
    \hfill
    \begin{subfigure}{0.19\linewidth}
        \includegraphics[width=\textwidth, trim=0 0 0 95, clip]{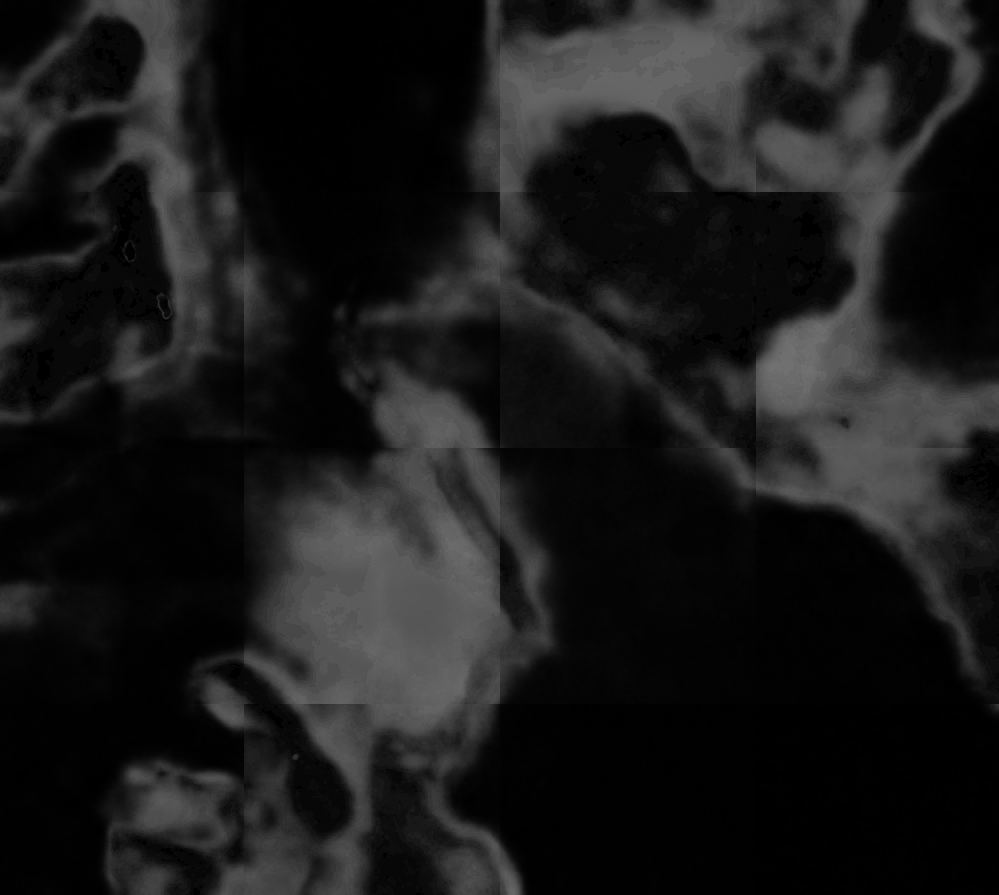}
    \end{subfigure}
    \hfill
    \begin{subfigure}{0.19\linewidth}
        \includegraphics[width=\textwidth, trim=0 0 0 95, clip]{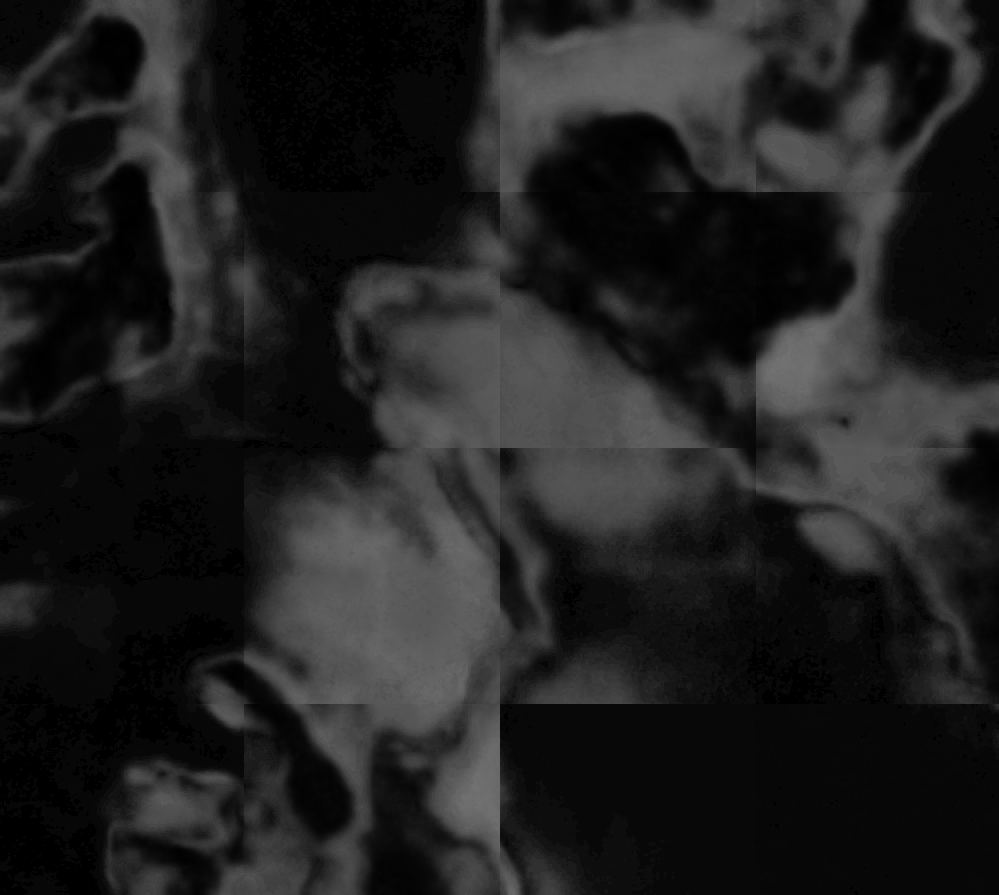}
    \end{subfigure}
    \hfill
    \begin{subfigure}{0.19\linewidth}
        \includegraphics[width=\textwidth, trim=0 0 0 95, clip]{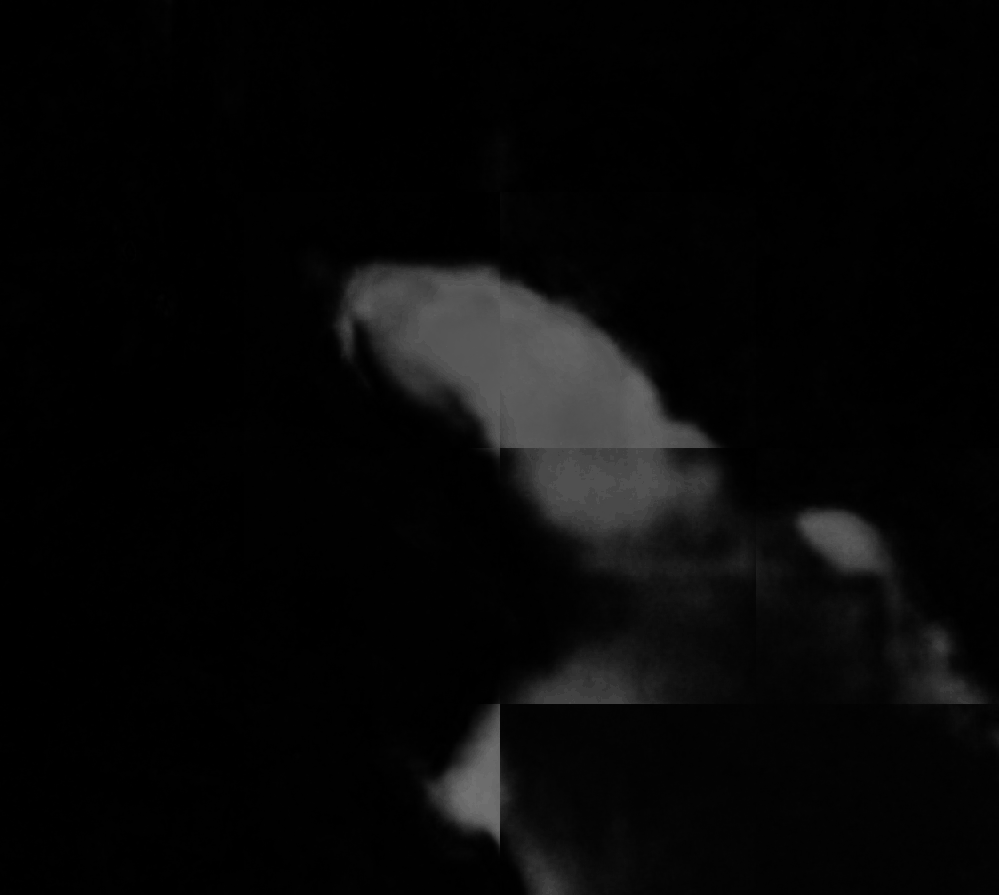}
    \end{subfigure}
    \vspace{7pt}

    \begin{subfigure}{0.19\linewidth}
        \includegraphics[width=\textwidth, trim=0 0 0 95, clip]{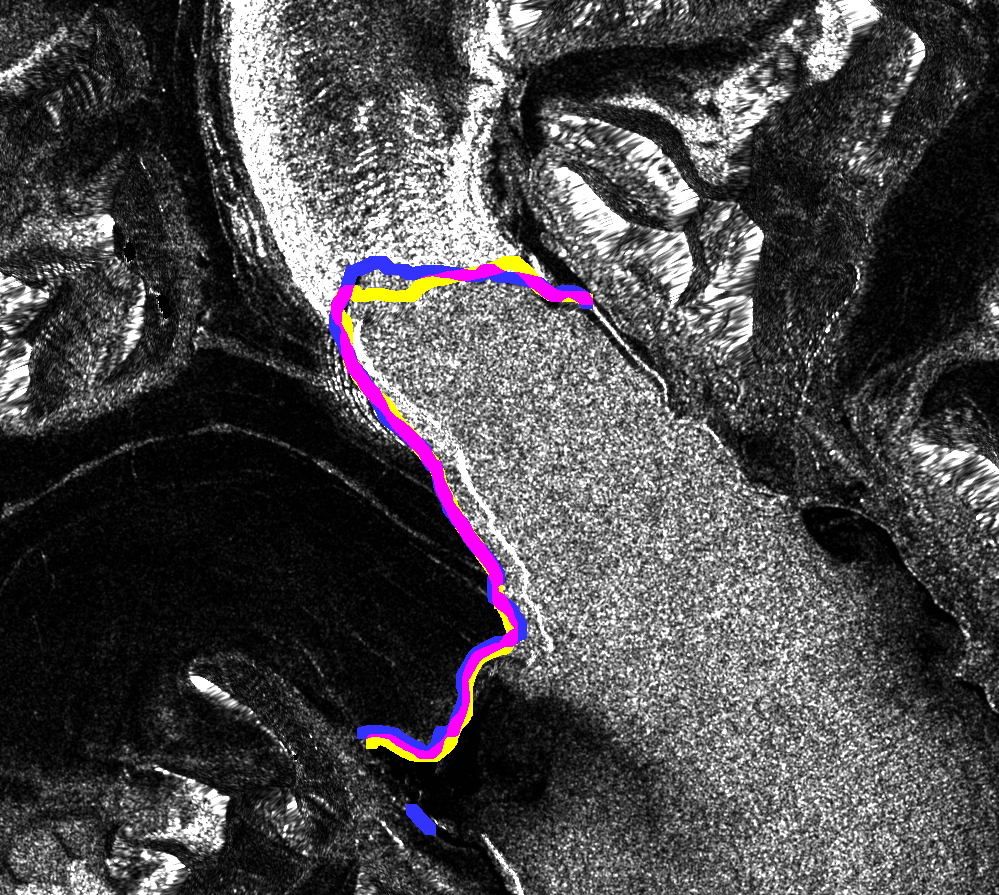}
    \end{subfigure}
    \hfill
    \begin{subfigure}{0.19\linewidth}
        \includegraphics[width=\textwidth, trim=0 0 0 95, clip]{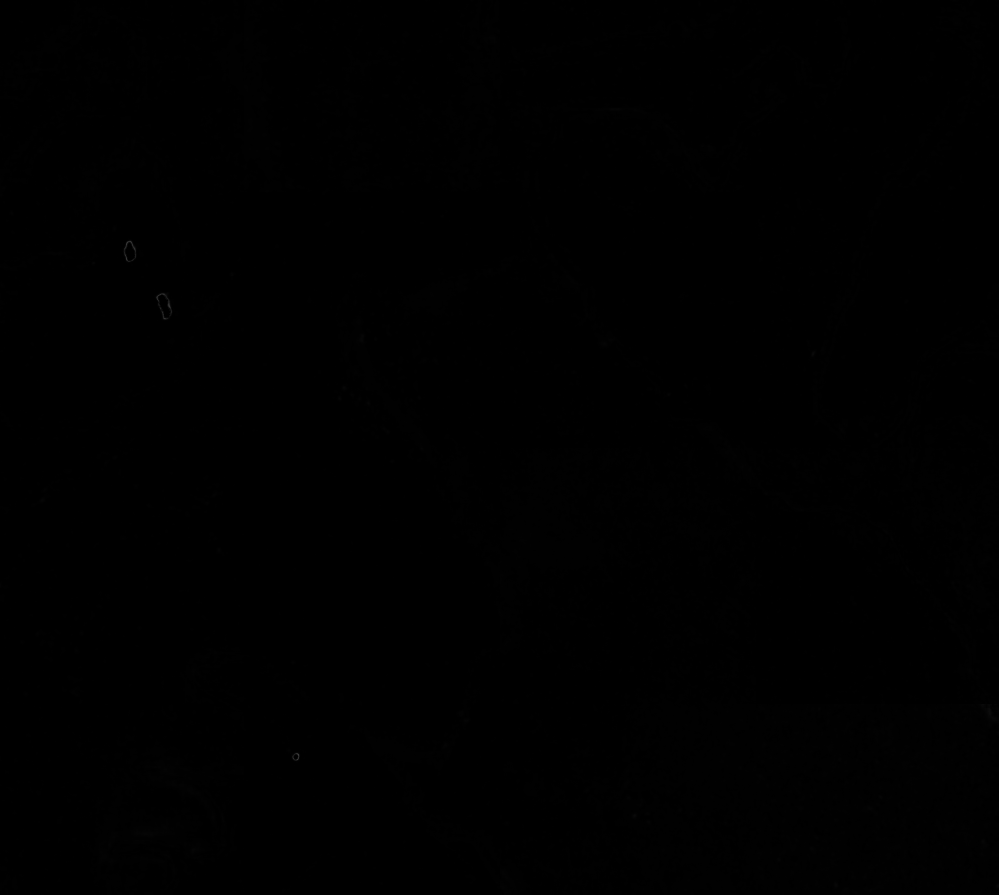}
    \end{subfigure}
    \hfill
    \begin{subfigure}{0.19\linewidth}
        \includegraphics[width=\textwidth, trim=0 0 0 95, clip]{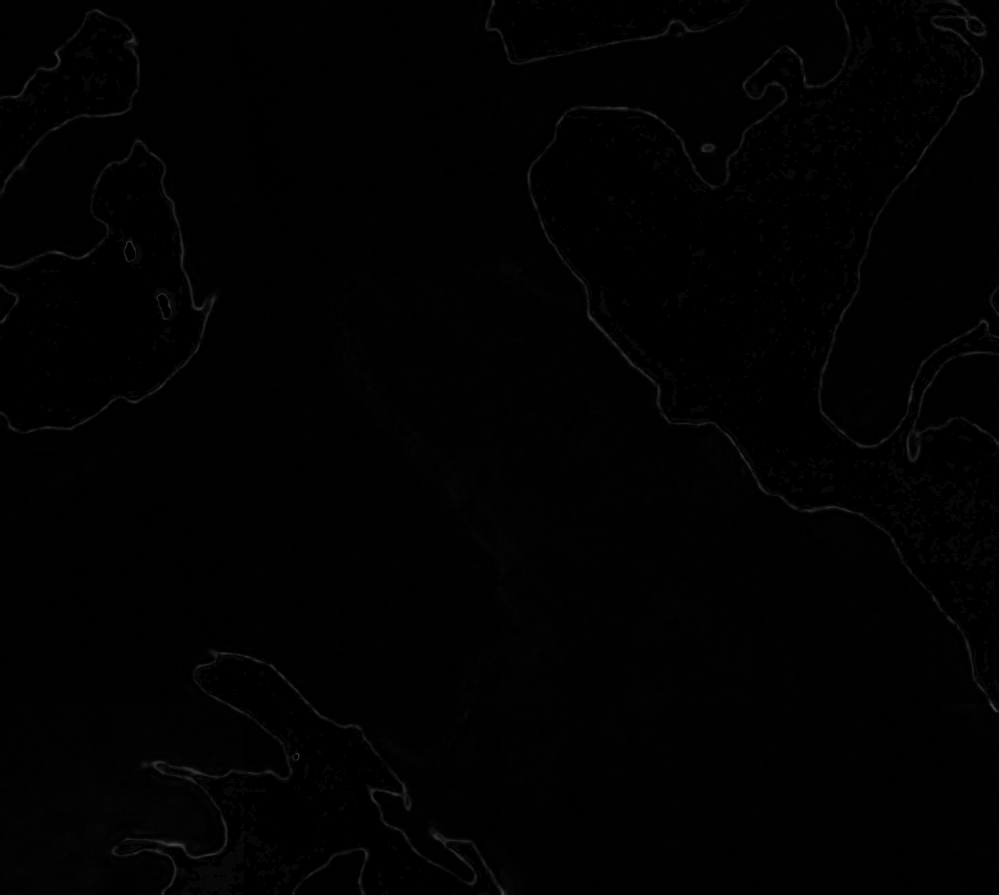}
    \end{subfigure}
    \hfill
    \begin{subfigure}{0.19\linewidth}
        \includegraphics[width=\textwidth, trim=0 0 0 95, clip]{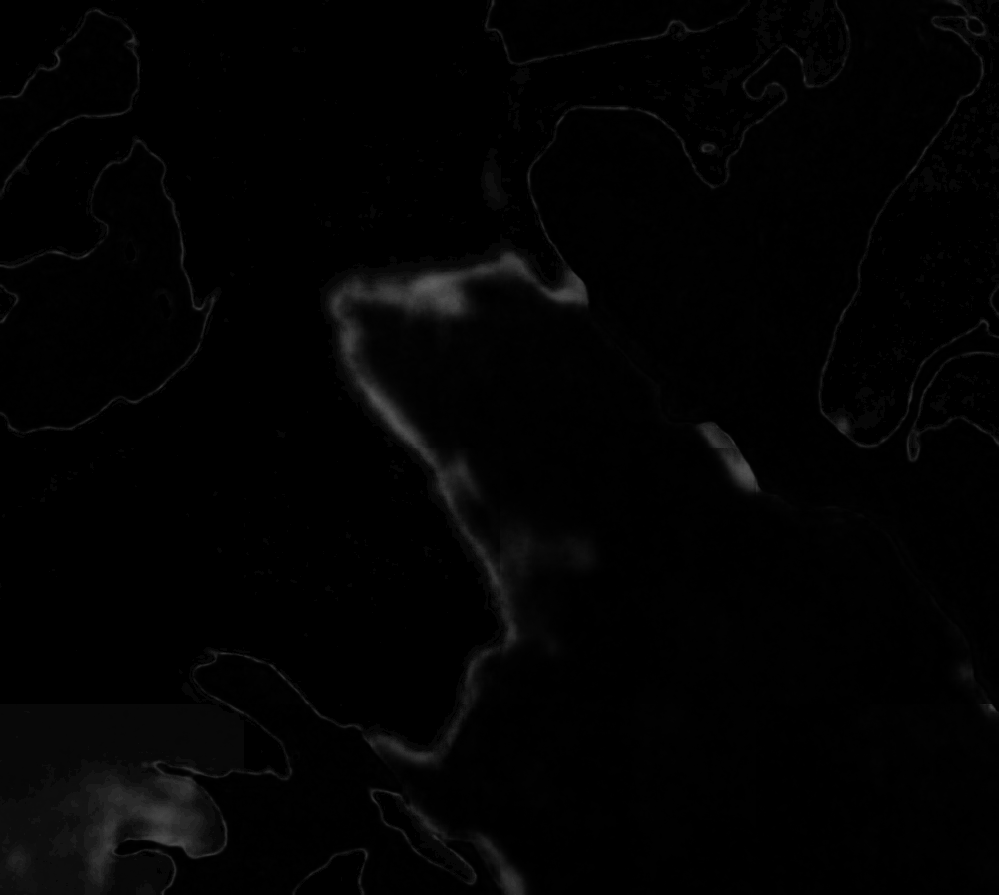}
    \end{subfigure}
    \hfill
    \begin{subfigure}{0.19\linewidth}
        \includegraphics[width=\textwidth, trim=0 0 0 95, clip]{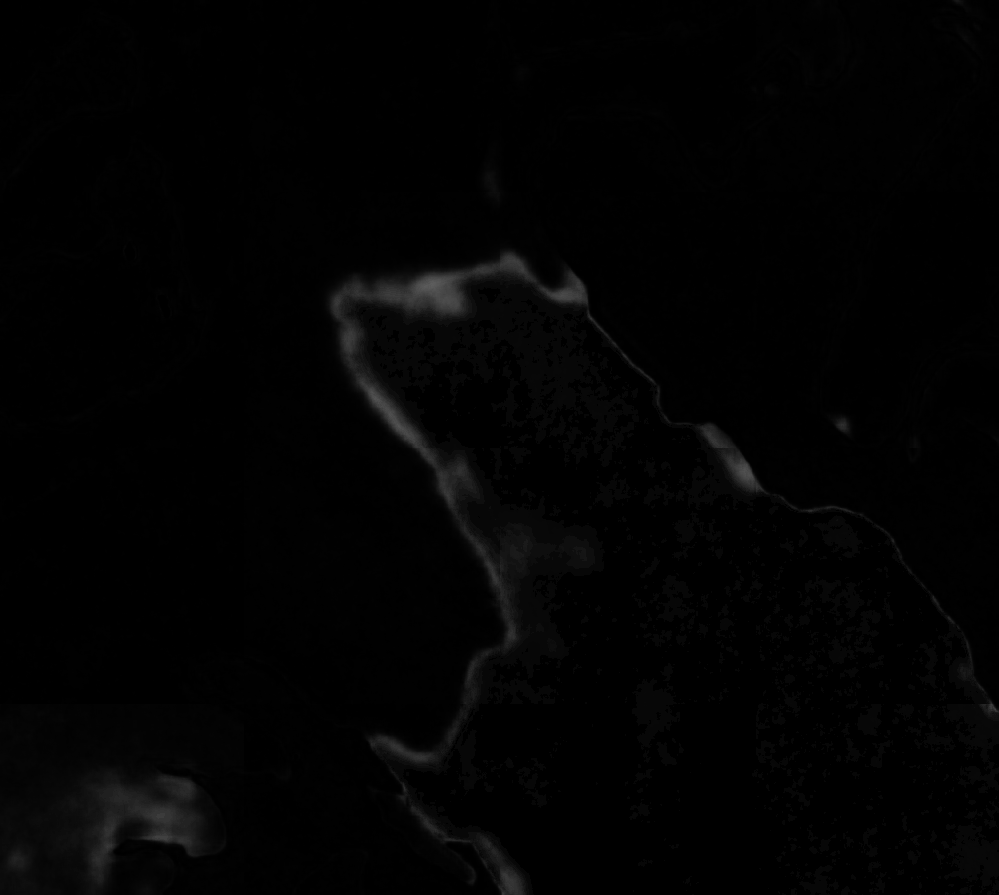}
    \end{subfigure}
    \caption{Uncertainty estimates for ensemble predictions of the baseline experiment (top row) and the final experiment (bottom row). Predicted fronts, ground truth fronts, and their overlap are shown in \fboxsep=1pt\colorbox{yellow!100}{yellow}, \fboxsep=1pt\colorbox{blue!100}{\color{white}blue}, and \fboxsep=1pt\colorbox{magenta!100}{\color{white}pink}, respectively. The calving fronts are dilated for visualization purposes. From left to right, panels show the ensemble prediction and the associated uncertainties for the NA, rock, glacier, and ocean classes.}
    \label{fig:uncertainty}
\end{figure*}

Finally, the ensemble of the final experiment achieves an \ac{mde} of \SI{68.7}{\metre} and an \ac{iou} of $81.1$, with only the NA \ac{iou} falling below $97.7$.
The lower \ac{iou} for the NA class may result from the small size of NA regions in the Svalbard dataset, which can cause issues with the patch embedding of the SwinV2 encoder.
Nevertheless, the segmentation performance for NA regions is irrelevant, as these areas correspond to masked pixels defined by radar pre-processing, with their locations known a priori for each image.
Figure~\ref{fig:vis} provides visualizations of the final calving fronts predicted by the ensemble for all five test set glaciers.
Figure~\ref{fig:uncertainty} shows how class-wise uncertainties differ between an ensemble of retrained baseline models and retrained models from the final experiment. 
Overall, uncertainties decrease when the three proposed methodological advancements are incorporated. 
While the baseline ensemble exhibits large regions of high uncertainty for the rock, glacier, and ocean classes, the ensemble from the final experiment shows elevated uncertainty only in areas immediately adjacent to the calving front and the coastline for the ocean and glacier classes. This indicates that the ice mélange-covered area is classified with high certainty. 
Almost no uncertainty remains for the rock class.

\section{Limitations and Future Work}
The findings of this study are limited to the test case, namely \ac{s1} images of the Svalbard archipelago. Application to other regions or sensors remains to be evaluated. A potential direction for future research is continual learning, which would allow the model to progressively incorporate new data as it becomes available. This approach would avoid retraining the model from scratch and enable adaptation to new study sites, sensors or acquisition settings while maintaining performance on previously learned data.

\section{Conclusion}
This study provides a transferable framework for applying deep learning–based calving front segmentation to novel study sites in a real-world setting. 
By applying a few-shot domain adaptation strategy, integrating spatial prior information, and incorporating summer reference images in the time series, the framework reduces the \acl{mde} of the state-of-the-art model on the Svalbard test site from \SI{1131.6}{\metre} to \SI{68.7}{\metre} without modifying the architecture. 
The results indicate that the proposed framework enables robust generalization and supports further expansion towards large-scale and potentially global deployment.

\section{Code and Data Availability}
The code is publicly available on GitHub\footnote{\url{https://github.com/ki7077/Real-World-Tyrion}}. 
The CaFFe benchmark dataset is already publicly available at PANGAEA\footnote{\url{https://doi.pangaea.de/10.1594/PANGAEA.940950}} and the new Svalbard dataset is publicly available at Zenodo\footnote{\url{https://zenodo.org/records/18196149}}.

\section{Author contribution}
Marcel Dreier, Nora Gourmelon: Conceptualization, Methodology, Software, Project administration, Writing - original draft preparation. \\
Dakota Pyles: Data curation, Writing - review \& editing. \\
Thorsten Seehaus, Matthias H. Braun, Andreas Maier, Vincent Christlein: Supervision, Writing - review \& editing. 

\section{Acknowledgments}
This research was funded by the Bayerisches Staatsministerium für Wissenschaft und Kunst within the Elite Network Bavaria with the Int. Doct. Program ``Measuring and Modelling Mountain Glaciers in a Changing Climate'' (IDP M3OCCA) as well as the German Research Foundation (DFG) project ``Large-scale Automatic Calving Front Segmentation and Frontal Ablation Analysis of Arctic Glaciers using Synthetic-Aperture Radar Image Sequences (LASSI)'' (Project number: 512625584), and the project ``PAGE'' within the DFG Emmy-Noether-Programme (DFG – SE3091/3-1; DFG – CH2080/5-1; DFG – SE3091/4-1).
The authors gratefully acknowledge the scientific support and HPC resources provided by the Erlangen National High Performance Computing Center (NHR@FAU) of the Friedrich-Alexander-Universität Erlangen-Nürnberg (FAU) under the NHR projects b110dc and b194dc. NHR funding is provided by federal and Bavarian state authorities. NHR@FAU hardware is partially funded by the DFG – 440719683.
The author team acknowledges the provision of satellite data under various AOs from respective space agencies (DLR, ESA, JAXA, CSA).
Map data copyrighted OpenStreetMap contributors and available from OpenStreetMap\footnote{\url{https://www.openstreetmap.org}}.

\bibliographystyle{IEEEbib}
\bibliography{refs}

\end{document}